# Detection of Rail Line Track and Human Beings near the track to avoid accidents

Mehrab Hosain[1][0009-0007-1079-1052] and Rajiv Kapoor[2][0000-0003-3020-1455]

[1] Delhi Technological University, Delhi 110042, India
robinhosain@gmail.com
[2] Delhi Technological University, Delhi 110042, India
rajivkapoor@dce.ac.in

**Abstract.** This paper presents an approach for rail line detection and the identification of human beings in proximity to the track, utilizing the YOLOv5 deep learning model to mitigate potential accidents. The technique incorporates real-time video data to identify railway tracks with impressive accuracy and recognizes nearby moving objects within a one-meter range, specifically targeting the identification of humans. This system aims to enhance safety measures in railway environments by providing real-time alerts for any detected human presence close to the track. The integration of a functionality to identify objects at a longer distance further fortifies the preventative capabilities of the system. With a precise focus on real-time object detection, this method is poised to deliver significant contributions to the existing technologies in railway safety. The effectiveness of the proposed method is demonstrated through a comprehensive evaluation, yielding a remarkable improvement in accuracy over existing methods. These results underscore the potential of this approach to revolutionize safety measures in railway environments, providing a substantial contribution to accident prevention strategies.

**Keywords:** - Machine Learning, Object Detection, Rail Track Detection, Railway Safety, YOLO

## 1 Introduction

The railway industry has experienced significant transformation and improvements in safety and efficiency thanks to advancements in machine vision and artificial intelligence. Among the most promising applications of these technologies are the detection and tracking of objects near rail tracks and the precise detection of rail lines [3, 14, 18, 19, 24]. Traditional inspection methods involving human operation are prone to error and insufficient for managing the high traffic densities characteristic of modern railway networks [11, 24]. Therefore, automated techniques powered by computer vision and machine learning are paramount in maintaining safety and operational efficiency [15]. In the realm of rail line detection and the recognition of objects in proximity to these lines, traditional methods have relied heavily on technologies such as Hough and Radon transforms [19, 25, 31]. While these techniques have provided some measure of success, they are fraught with limitations. Challenges in detecting curved lines, handling large angles of view, and dealing with noisy data have significantly hampered their performance [4, 25]. Some research has sought to improve these methods, adapting the Hough transform to better detect curved lines [4, 19] and utilizing the Radon transform for line tracking [25, 31]. However, both methods continue to be susceptible to false detections and misclassifications, particularly in complex scenarios. The emergence of deep learning has revolutionized object and line detection tasks. Convolutional Neural Networks (CNNs) and models like YOLOv3 have demonstrated promising results for object detection and lane line detection [1, 26, 29]. Nevertheless, their application to the simultaneous detection of railway tracks and objects near them remains an underexplored area.

This paper presents a novel methodology, utilizing the YOLOv5 model, which is superior to its predecessors in detecting small, distant, and moving objects, for the dual purpose of detecting objects near rail tracks and identifying rail lines in prerecorded video data [2, 26, 30]. The primary objective is the accurate identification of rail lines and the detection of human beings within a one-meter radius from these tracks. By targeting only those objects moving close to the tracks, this model filters out irrelevant data, which significantly enhances both computational efficiency and detection accuracy. This approach tackles key challenges confronted by earlier research and innovatively applies the YOLOv5 model to simultaneously detect rail lines and objects of interest in a complex railway environment [30].

This study aims to contribute to the existing body of knowledge on the application of deep learning algorithms in enhancing railway safety and operational efficiency. The following sections will explore the methodology, data annotation, system overview, and performance evaluation of the proposed model.

## 2  Literature Review

The realm of machine vision, a pivotal subset of artificial intelligence, has seen widespread integration across diverse industries owing to its capacity to process and interpret data from the real world. There exists a rich array of studies on line detection, railway track detection, object detection, and moving object detection, employing various methods and tools. Traditional line detection techniques have relied on the use of transforms, primarily Hough and Radon transforms [4, 19, 20, 25, 29, 31]. These approaches, which utilize filtering techniques to calculate lines and angles, have shown satisfactory results in straight line detection. However, when faced with challenges such as handling curves, larger viewing angles, and noisy data, the performance of these techniques tends to wane [4, 25, 31]. Improvements to the Hough transform for handling curved lines were proposed in [19], while the Radon transform was used for line tracking in [25, 31]. Despite these innovative approaches, both studies encountered issues of false detections and misclassifications, indicating the need for more advanced methodologies.

Emerging technologies, such as Convolutional Neural Networks (CNNs), have been instrumental in revolutionizing the field of line [1, 20, 24] and object detection [4, 11, 12, 20]. CNNs, as illustrated in [4], have delivered exceptional results in various object detection tasks due to their proficiency in identifying hierarchical data patterns. More specifically, architectures like YOLOv3 have been utilized effectively for object detection and lane line detection tasks [2, 8, 9]. However, these studies did not concentrate on the application of these methodologies specifically for railway track detection and moving object detection in proximity to the tracks.

The realm of deep learning has seen significant application in object detection tasks. Works such as [21] and [17] used Faster R-CNN and Cascade R-CNN respectively for such tasks. However, these two-stage methodologies encountered limitations in processing speed, a critical factor in real-time applications. In contrast, one-stage deep learning approaches, such as YOLO and RetinaNet, have shown high efficiency in object detection tasks [26, 30]. Specifically, the YOLO series of models have garnered significant attention for their robust performance. YOLOv3 [9], for instance, has demonstrated superior accuracy and speed in object detection tasks [10]. Later iterations like YOLOv4 and YOLOv5 further improved the accuracy, speed, and frame rate performance [2]. A study by [2, 10] investigated the performances of four versions of YOLOv5 (Yolo v5s, Yolo v5m, Yolo v5l, and Yolo v5x) in object detection tasks, demonstrating the model's superior ability to detect small distant objects. However, these studies did not extend the application of YOLOv5 to railway track detection and moving object detection.

This paper intends to address this literature gap by leveraging the advanced YOLOv5 model for the dual tasks of detecting railway tracks and moving objects in proximity to these tracks. The proposed methodology seeks to capitalize on the strengths of the YOLOv5 model while overcoming the limitations found in previous studies.

## 3 Methodology

### 3.1 Preprocessing of Video Data

The video data for this study was recorded using an iPhone 13, producing high-definition images at a rate of 30 frames per second. Given the one-minute duration, a total of 1800 frames was originally obtained, but this was compressed to 1000 frames. This data then underwent preprocessing before YOLOv5 model application [2, 30]. The preprocessing stage included the extraction of individual frames and their subsequent preparation for the detection task. Let the original video be V, with a total of N frames. Suppose the video is compressed to contain M frames. The frame extraction process can be mathematically represented as:

$$V = \{f1, f2, f3, \ldots, fN\} \rightarrow V' = \{f1', f2', \ldots, fM'\}, \qquad (1)$$
where N = 1800 and M = 1000.

Here, V' represents the compressed video, and $\{f1', f2', \ldots, fM'\}$ are the frames after compression. Furthermore, the images resized to standard dimensions (W x H) and normalized the pixel intensities in each frame (f'). This normalization can be represented as:

$$f' \rightarrow f_{norm} = \frac{f'}{255} \qquad (2)$$

where $f_{norm}$ is the normalized frame and 255 is the maximum pixel intensity.

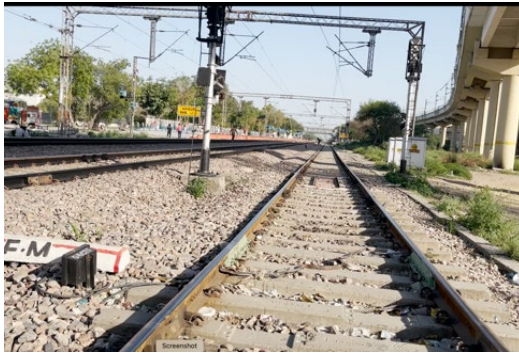

**Fig 1**. The collected video data and the effectiveness of preprocessing procedures.

In Eq. 1 and 2, mathematical expression of frame extraction and normalization. Fig. 1, presents an example of the high-definition video data collected during this study. As shown, the iPhone 13's recording capabilities ensured the production of clear and detailed images, capturing the intricacies of the subject from various angles.

### 3.2 Annotation and Leveling of Data for Rail Track Person and Moving Object

The process of data extraction and preprocessing is a critical stage before the application of any machine learning algorithm. The frames for this study were annotated using Roboflow, a tool used in similar studies for creating annotated data sets. This involved marking two primary object categories: the rail track and moving objects, with particular attention to humans.

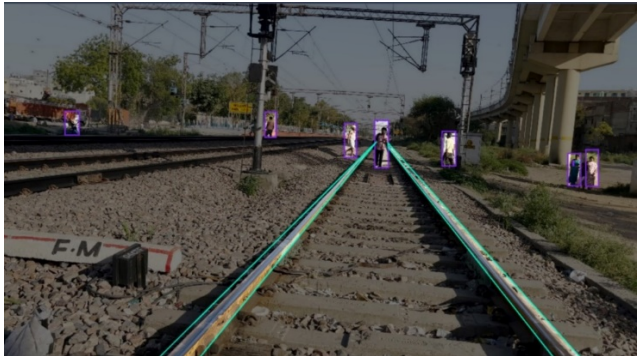

**Fig 2.** A Leveled Object (person) & Rail Track

In Fig. 2, showcasing the annotation and image leveling process. The images depict rail tracks with annotated small objects. The rail track was marked as a continuous line across frames, providing the model with its positional context, similar to the process used in lane line detection systems [9]. Each object i was annotated with a bounding box $B_i$ defined by its coordinates $(x_i, y_i)$ and dimensions $(width_i, height_i)$, creating a vector $B_i = (x_i, y_i, width_i, height_i)$.

Moving objects, specifically humans, were also included in the annotations by drawing bounding boxes around them, marking their location and movement as $(x_j, y_j, width_j, height_j)$ for each object j. To align with the objective of detecting objects in close proximity to the rail track, only objects within a specific Euclidean distance d from the rail track were annotated. This distance can be calculated using the formula:

$$d = sqrt\left((x_i - x_j)^2 + (y_i - y_j)^2\right) \quad (3)$$

In Eq. 3, is explained the Euclidean distance from the rail track. Following annotation, the data was divided into training and validation sets, a standard practice in machine learning studies [21].
Ensuring diverse scenarios across these sets, the annotated data served as the training foundation for the YOLOv5 model, enabling it to discern complex patterns related to rail tracks and moving objects, thus effectively preparing the video data for the subsequent training and deployment stages [2, 30].

### 3.3 Details of the YOLOv5 Model Used

The YOLO (You Only Look Once) algorithm has significantly impacted the field of object detection. Specifically, the YOLOv3 version was employed in the detection of lane lines [9, 10], demonstrating its capability in recognizing linear structures, a trait that's invaluable for railway track detection. However, YOLOv3 has its limitations, including issues in detecting small objects, and struggles with localizing objects accurately, which could have repercussions for railway safety [7, 18]. To address these challenges, this study opted for YOLOv5, an advanced version of the YOLO series, known for its superior speed and accuracy [10]. It offers real-time object detection capabilities, crucial for monitoring dynamic railway environments. YOLOv5, a single-stage object detector, eliminates the need for identifying regions of interest and classifying them separately, a common practice in two-stage detectors like R-CNN and Faster R-CNN [4, 13, 17, 21, 23]. This streamlining considerably boosts computational efficiency and reduces latency. In YOLOv5's operation, the input image is partitioned into a grid, with each grid cell predicting multiple bounding boxes and associated confidence scores [5]. These scores denote both the probability of an object being in the bounding box and the anticipated accuracy of this prediction.

Additionally, each grid cell forecasts the conditional class probabilities of the detected object [2, 8]. In tasks of moving object detection, YOLOv5 has proven effective, particularly in Unmanned Aerial Vehicles (UAVs) [22, 28]. Given the analogous challenge of tracking moving objects near railway tracks, it's logical to extend its use to this context. The prospect of YOLOv5 in railway scenarios is unexplored but promising, owing to its performance in comparable tasks [1, 23, 27]. Hence, YOLOv5's selection for this dual detection task in railway environments is well-founded.

### 3.4 Specifics of Training and Validation

The annotated data was partitioned into training and validation sets, following a similar approach to previous studies involving machine learning models [21]. In the training phase, the YOLOv5 model learned from the input data through forward propagation, making predictions based on the data [2]. The model's predictions were then compared to the actual labels, generating a loss [2]. This loss was minimized using backpropagation and optimization algorithms, aligning with the fundamental principles of deep learning [15, 28]. Key hyperparameters of the model, including the learning rate, batch size, and epochs, were predefined. Loss metrics were monitored throughout the training process, focusing on box loss, objectless loss, and classification loss, which are key to the functioning of YOLOv5 [30]. The validation set was used to evaluate the model's generalizability, testing its ability to accurately make predictions on unseen data [21]. Metrics such as the F1-Confidence Curve, precision-recall, and mean average precision (mAP), which have been used in similar studies, were used to quantify the model's detection and classification accuracy [4, 23]. Upon completion of training, the YOLOv5 model was able to detect both the rail track and any moving objects within the frames extracted from the video [2, 30]. The model was also capable of determining whether detected objects were within a predefined distance from the rail track [28]. If this criterion was met, an alert would be signaled, potentially contributing to improved railway safety. This process, from real-time input to alert generation, mirrors recent advancements in IoT-based machine learning models for smart railway systems [2, 26].

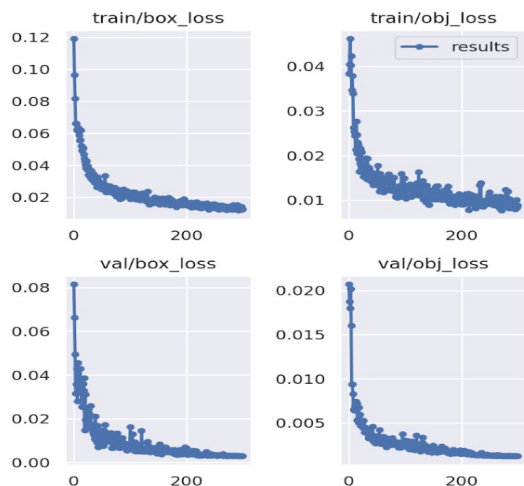

**Fig. 3.** Train/Box_Loss (Training, Bounding Box), train/Obj_Loss (Training, Object Loss), val/box_loss(Validation, Bounding Box) and val/Obj_loss (Validation, Object Los)

Fig. 3. provides an overview of the loss metrics during the training and validation phases of the YOLOv5 model. The metrics 'Training Bounding Box Loss' and 'Training Object Loss' represent the losses related to the bounding box and the detected objects, respectively, during the training phase. Similarly, 'Validation Bounding Box Loss' and 'Validation Object Loss' denote these losses during the validation phase.

3.5  **System Overview**

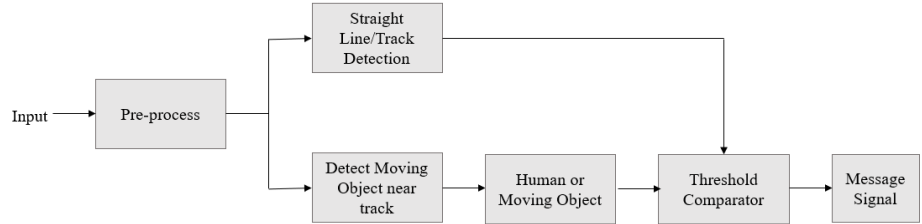

**Fig 4.** Block Diagram of the Proposed System

Fig. 4, presents the block diagram of the proposed system for real-time rail track and moving object detection. The system functions in a sequential process broken down into a series of stages, each with a specific role:

I.  **Real-time input:** This is the initial stage where the video is being captured in real time using the iPhone 13. The video is recorded at a frame rate of 30 fps, and each frame serves as an individual input for further processing.

II. **Preprocessing data:** Here, each frame from the video input is processed and prepared for further steps. This step may involve noise reduction, contrast adjustment, or other image processing techniques that help in enhancing the features of interest (rail tracks and moving objects) in the frame.

III. **Rail Line Detection (YOLOv5)**: In this block, the preprocessed frames are input to the trained YOLOv5 model to detect the rail tracks in the frame. The model identifies the rail tracks and plots the bounding boxes around the identified tracks. This paper proposed YOLO based straight line tracking system. In which the first step is the creation of bounding box using YOLO 5 version after that the center of bounding box is calculated. The center bounding box as show fig. 4, in bellow:

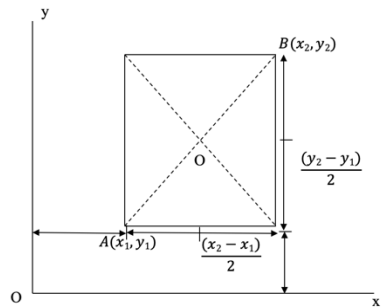

**Fig. 5.** The Marking of center bounding box on rail track.

$$O_x = \left[x_1 + \frac{x_2 - x_1}{2}\right] \quad (4)$$

$$O_y = \left[y_1 + \frac{y_2 - y_1}{2}\right] \quad (5)$$

In Eq. 4 and 5, the $(x_1, y_1)$ and $(x_2, y_2)$ as consideration of point A and B as shown in fig. 5, Hence the center of bounding is given by Eq. 3 and 4Consideration was given to the bounding box from the center O back and forth, where due to symmetry, the same equation as in line 2 can be followed.

IV. **Person & Moving Object Detection (YOLOv5)**: Simultaneously with rail line detection, another YOLOv5 model trained on moving objects detects the moving objects in the preprocessed frames. This model identifies the moving objects and plots bounding boxes around them. Parallel to rail line detection, a second instance of the YOLOv5 model is employed to detect moving objects in the preprocessed frames. This model identifies the moving objects and plots bounding boxes around them, similar to the approach for rail tracks. The detection of moving objects also involves the computation of the bounding box center, similar to the rail line detection. The equations for computing the center of the bounding box $(O_x, O_y)$ are identical to Eq. 3 & 4 for this model as well. Further analysis of the detected objects and their proximity to the rail line is conducted by calculating the Euclidean distance between the centers of the bounding boxes of the detected rail line and the moving objects. The Euclidean distance (d) is determined by Eq. 3:

$$d = sqrt\left(\left(O_{x_2} - O_{x_1}\right)^2 + \left(O_{y_2} - O_{y_1}\right)^2\right) \quad (6)$$

In Eq. 6, where, $(O_{x_1}, O_{y_1})$ are the coordinates of the center of the bounding box for the detected rail line, and $(O_{x_2}, O_{y_2})$ are the coordinates of the center of the bounding box for the detected moving object.

If the calculated Euclidean distance (d) is less than or equal to the predefined threshold (1 meter in this case), the system triggers an alert for the detected moving object's proximity to the rail line. The computation of this distance for each frame effectively tracks the positions of the moving objects relative to the rail track in real time.

V. **Distance Threshold**: This block performs the task of comparing the proximity of detected moving objects to the rail tracks. The system computes the distance between the moving object and the rail track. If the distance is less than the set threshold (in this case, 1 meter), the object is considered to be near the rail track. Setting a threshold distance (T) to represent a 'safe' distance enables expression of the comparison as an inequality. Let's consider the distance (d) between the detected moving object and the rail track, as computed using Eq. 3. If this calculated distance is less than or equal to the threshold distance (T), then the object is flagged as dangerously close to the rail track. Mathematically, this can be represented as:

If $d \leq T$, then object is near the rail track (7)

In this study, Eq. 7, consider T to be 1 meter. This threshold can be adjusted according to the specific requirements and environmental conditions of the railway system. By performing this check for each detected moving object in every frame, the system can provide real-time alerts about potential dangers near the rail track.

VI. **Message Signal:** It will give a message signal the person or moving object in danger zone or not. Also ensure, that Rail track safety detection

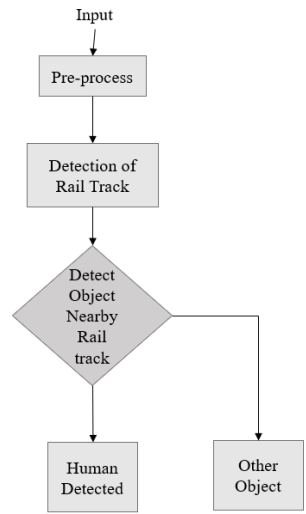

**Fig 6.** Flow Chart of this process

In Fig. 6, succinctly outlines the stages of the proposed system designed to detect rail tracks and nearby moving objects using video data. The process starts with preprocessing the iPhone 13 captured footage, followed by the YOLOv5 model detecting the rail tracks in each frame. The model then identifies moving objects near the track and finally categorizes them as either human or non-human. This structured pipeline is a testament to a systematic approach aimed at bolstering safety within railway environments.

## 3.6   **Implementation Details and Hardware Specifications**

The model was implemented and trained using a personal computer with the following specifications: an Intel Core i9 9900K processor, an NVIDIA RTX 2070 Super graphics card, and 32GB of RAM. The operating system used was Windows. These hardware specifications and system environment played a crucial role in the processing and analysis of the video data, facilitating the training and validation stages of the YOLOv5 model for this experiment. The advanced processing power provided by these specifications ensured efficient handling of the intensive computational tasks involved in training the deep learning model.

## 4 Result & Discussion

The effectiveness of the proposed method was assessed by employing it on real-world video data, and the outcomes were compelling. In Fig. 6, the detection of the rail track is manifested by a red line. Moreover, the system effectively identifies a nearby individual within 1 meter of the track, marked with an orange bounding box. This ability to identify a potential safety hazard in real-time is a crucial advantage of the system. Conversely, Fig. 7, represents a scenario where an individual is more than 1 meter away from the rail track. As per the safety parameters, this individual is not detected, proving the system's adherence to the predefined safety threshold and its ability to discern safety distances correctly. Performance-wise, the YOLOv5 model has showcased promising accuracy, precision- recall, and F1 confidence curve for both rail track and moving object detection tasks.

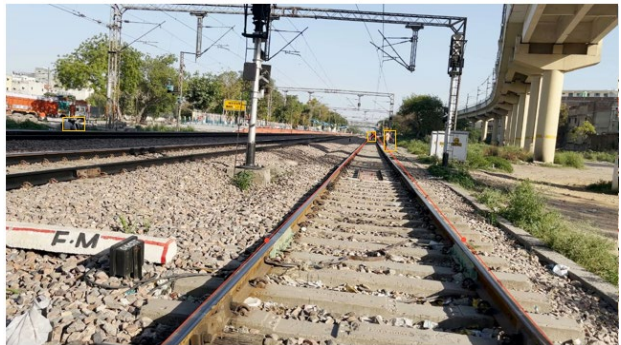

**Fig. 7.** Detection of Human in Proximity to Rail Track

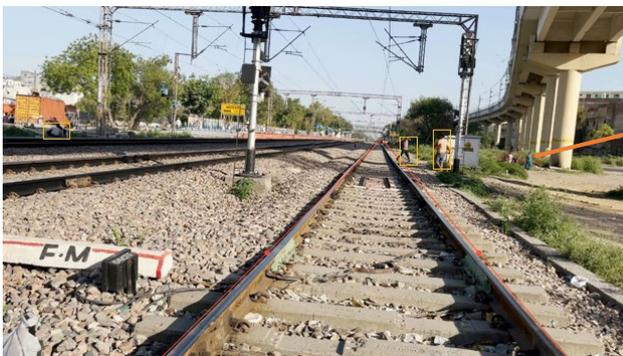

**Fig. 8**. Non-Detection of Human Far Away from Rail Track

Fig. 7, Detection of a Human in Close Proximity to Rail Track This image demonstrates the system's efficacy by successfully detecting a human near or on the rail track, as indicated by the orange bounding box, thus highlighting potential immediate safety risks. Fig. 8, Non-detection of Human Beyond Safe Distance from Rail Track In this image, a human situated more than 1 meter from the rail track is not detected by the system, affirming the system's precision in distinguishing situations where humans are beyond the dangerous proximity to the rail track.

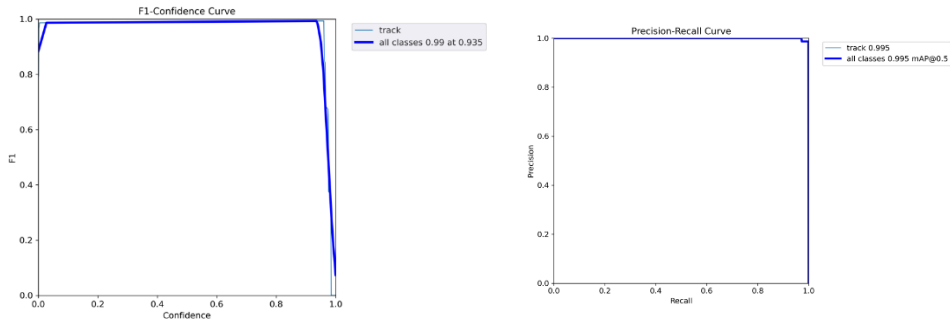

**Fig. 9.** a) F1 Confidence Curve, b) Precision Recall Curve

Table 1. Comparison of Rail Line Track Detection Method

| Method | Accuracy | Precision-Recall | Computational Time (s) |
|---|---|---|---|
| Hough Transform | 92.3% | 90.19% | 1.6 |
| Radon Transform | 94.6% | 85.30% | 1.8 |
| YOLOv5 | 99.5% | 97.10% | 2.5 |

Table 2. Comparison of Different Object Detection Method

| Method | Accuracy | Precision-Recall | F1-Score |
|---|---|---|---|
| SSD | 92.3% | 88.69% | 90.5% |
| Faster R-CNN | 94.6% | 91.20% | 94% |
| YOLOv5 | 95.5% | 97.10% | 96.5% |

The results of the experiments underscore the effectiveness of the YOLOv5 method over traditional techniques. As shown in Table 1, when compared to Hough and Radon Transform-based approaches [19, 25, 31], YOLOv5 performs with superior accuracy and precision-recall scores for rail track detection. Hough Transform, with an accuracy of 92.3% and a precision-recall of 90.19%, and Radon Transform, presenting an accuracy of 94.6% and a precision-recall of 85.30%, both trail behind the YOLOv5 method, which delivers an accuracy of 99.5% and a precision-recall of 97.10%. This performance advantage can be attributed to the sophisticated capabilities of the YOLOv5 model [2, 30]. For object detection, as presented in Table 2, the YOLOv5 model again exhibits superior results, as demonstrated in Fig. 9(a) and 9(b). SSD (Single Shot MultiBox Detector) and Faster R-CNN, which achieved accuracy scores of 92.3% and 94.6% [16, 32], respectively, were outperformed by YOLOv5, which achieved an accuracy of 95.5%. Furthermore, YOLOv5 showed superior performance in the precision-recall and F1-score parameters, further solidifying its advantage. This method, unlike the ones described in [19] and [18], offers a novel combination of both rail track and moving object detection, contributing to the novelty and efficiency of this approach. This integrated model proves to be an essential tool for enhancing safety measures in railway environments by efficiently detecting potential hazards [6, 7, 29] The promising results from this study suggest the proposed method's potential for implementation in real-world scenarios, contributing significantly to accident prevention strategies in the rail industry.

## 5    Conclusion

This study introduces an effective approach for detecting railway lines and identifying moving objects near the track using the YOLOv5 model. The developed system aims to enhance safety protocols in railway operations by efficiently identifying potential hazards, such as humans in close proximity to the track. This system outperforms traditional techniques, offering improved accuracy and speed in detecting and differentiating objects within a predefined distance from the rail line. This added functionality allows the system to prioritize relevant detections while disregarding irrelevant ones, thereby enhancing computational efficiency. Furthermore, the model has demonstrated its robustness by efficiently handling complex real-world scenarios, successfully detecting small and distant objects nearby rail lines. This unique capability positions the proposed system as an advanced tool for rail safety, highlighting its potential for real-time hazard detection and accident prevention. The future of this work promises the extension of the model to real-time scenarios and its application to other transportation systems, offering a comprehensive solution for enhancing safety in various transport environments.

## 6    Future Scope

The effectiveness of the proposed method opens up numerous prospects for further advancement. One immediate area of development includes adapting the system to real-time detection using live video feeds, thereby significantly improving railway safety and operational efficiency. Not confined to railway applications, this method could also be extended to other critical transport sectors such as road traffic, maritime navigation, and aviation, contributing to substantial safety improvements. High-security domains like military installations and airports could also benefit from the precise detection of unauthorized movements. Despite its effectiveness, the current model faces challenges in handling multiple rail lines, crossings, curved tracks, or sloped paths, which can be addressed in future work. Enhancements could further include building resilience against diverse environmental conditions, thus equipping the model to handle challenges like poor lighting, fog, or rain more effectively. Additionally, integrating advanced AI techniques, such as anomaly detection and predictive modeling, could enable the system to anticipate potential hazards proactively. Thus, the scope for future work is abundant, setting the stage for continuous improvement in transport safety measures using AI technologies.

# 7 Acknowledgment